\documentclass{article}

\usepackage{agi-2010}
\usepackage{graphicx}
\usepackage{amsmath}
\usepackage{amssymb}
\usepackage{amsthm}
\usepackage{psfrag}
\usepackage{mathrsfs}
\usepackage{booktabs}
\usepackage{natbib}

\newcommand{\field}[1]{\mathbb{#1}}
\newcommand{\power}{\mathscr{P}}
\newcommand{\fs}[1]{\mathcal{#1}}

\newcommand{\nats}{\field{N}}
\newcommand{\reals}{\field{R}}

\newcommand{\define}{\equiv}

\newcommand{\comment}[1]{}

\newcommand{\g}[1]{\underline{#1}}

\newcommand{\finint}{\fs{Z}^\ast}   
\newcommand{\infint}{\fs{Z}^\infty} 
\newcommand{\salgebra}{\fs{F}}      
\newcommand{\prob}{\mathbf{P\!r}}   
\newcommand{\gen}{\mathbf{G}}  
\newcommand{\agent}{\mathbf{P}}     
\newcommand{\env}{\mathbf{Q}}       
\newcommand{\rew}{\mathbf{r}}       
\newcommand{\utility}{\mathbf{U}}   
\newcommand{\genutility}{\mathbf{G\!U}}
\newcommand{\predutility}{\mathbf{P\!U}}
\newcommand{\expect}{\mathbf{E}}    
\newcommand{\entropy}{\mathbf{H}}   
\newcommand{\kldiv}{\mathbf{KL}}    

\theoremstyle{plain}
\newtheorem{theorem}{Theorem}

\newtheorem{proposition}{Proposition}
\newtheorem{corollary}{Corollary}

\theoremstyle{definition}
\newtheorem{definition}{Definition}

\begin{document}

\bibliographystyle{plainnat}

\title{A conversion between utility and information}

\author{Pedro A. Ortega\\
        Department of Engineering\\
        University of Cambridge\\
        Cambridge CB2 1PZ, UK\\
        \texttt{peortega@dcc.uchile.cl}\\
        \And
        Daniel A. Braun\\
        Department of Engineering\\
        University of Cambridge\\
        Cambridge CB2 1PZ, UK\\
        \texttt{dab54@cam.ac.uk}\\
}


\maketitle

\begin{abstract}%
Rewards typically express desirabilities or preferences over a set of
alternatives. Here we propose that rewards can be defined for any probability
distribution based on three desiderata, namely that rewards should be
real-valued, additive and order-preserving, where the latter implies that more
probable events should also be more desirable. Our main result states that
rewards are then uniquely determined by the negative information content. To
analyze stochastic processes, we define the utility of a realization as its
reward rate. Under this interpretation, we show that the expected utility of a
stochastic process is its negative entropy rate. Furthermore, we apply our
results to analyze agent-environment interactions. We show that the expected
utility that will actually be achieved by the agent is given by the negative
cross-entropy from the input-output (I/O) distribution of the coupled
interaction system and the agent's I/O distribution. Thus, our results allow
for an information-theoretic interpretation of the notion of utility and the
characterization of agent-environment interactions in terms of entropy
dynamics.
\end{abstract}


\section{Introduction}
Purposeful behavior typically occurs when an agent exhibits specific
preferences over different states of the environment. Mathematically, these
preferences can be formalized by the concept of a utility function that assigns
a numerical value to each possible state such that states with higher utility
correspond to states that are more desirable \citep{Fishburn1982}. Behavior can
then be understood as the attempt to increase one's utility. Accordingly,
utility functions can be measured experimentally by observing an agent choosing
between different options, as this way its preferences are revealed.
Mathematical models of rational agency that are based on the notion of utility
have been widely applied in behavioral economics, biology and artificial
intelligence research \citep{RussellNorvig1995}. Typically, such rational agent
models assume a distinct reward signal (or cost) that an agent is explicitly
trying to optimize.

However, as an observer we might even attribute purposefulness to a system that
does not have an explicit reward signal, because the dynamics of the
system itself reveal a preference structure, namely the preference over all
possible paths through history. Since in most systems not all of the histories
are equally likely, we might say that some histories are more probable than
others because they are more desirable from the point of view of the system.
Similarly, if we regard all possible interactions between a system and its
environment, the behavior of the system can be conceived as a drive to generate
desirable histories. This imposes a conceptual link between the probability of
a history happening and the desirability of that history. In terms of agent
design, the intuitive rationale is that agents should act in a way such that
more desired histories are more probable. The same holds of course for the
environment. Consequently, a competition arises between the agent and the
environment, where both participants try to drive the dynamics of their
interactions to their respective desired histories. In the following we want to
show that this competition can be quantitatively assessed based on the entropy
dynamics that govern the interactions between agent and environment.

\section{Preliminaries}

We introduce the following notation. A set is denoted by a calligraphic
letter like $\fs{X}$ and consists of \emph{elements} or \emph{symbols}.
\emph{Strings} are finite concatenations of symbols and \emph{sequences} are
infinite concatenations. The empty string is denoted by $\epsilon$. $\fs{X}^n$
denotes the set of strings of length $n$ based on $\fs{X}$, and $\fs{X}^\ast
\define \bigcup_{n \geq 0} \fs{X}^n$ is the set of finite strings. Furthermore,
$\fs{X}^\infty \define \{x_1 x_2 \ldots | x_i \in \fs{X} \text{ for all }
i=1,2,\ldots \}$ is defined as the set of one-way infinite sequences based on
$\fs{X}$. For substrings, the following shorthand notation is used: a string
that runs from index $i$ to $k$ is written as $x_{i:k} \define x_i x_{i+1}
\ldots x_{k-1} x_k$. Similarly, $x_{\leq i}
\define x_1 x_2 \ldots x_i$ is a string starting from the first
index. By convention, $x_{i:j} \define \epsilon$ if $i>j$. All proofs can be
found in the appendix.

\section{Rewards}

In order to derive utility functions for stochastic processes over finite
alphabets, we construct a utility function from an auxiliary function that
measures the desirability of events, i.e. such that we can assign desirability
values to every finite interval in a realization of the process. We call this
auxiliary function the reward function. We impose three desiderata on a reward
function. First, we want rewards to be mappings from events to reals numbers
that indicate the degree of desirability of the events. Second, the reward of a
joint event should be obtained by summing up the reward of the sub-events. For
example, the ``reward of drinking coffee and eating a croissant'' should equal
``the reward of drinking coffee'' plus the ``reward of having a croissant given
the reward of drinking coffee''\footnote{Note that the additivity property does
not imply that the reward for two coffees is simply twice the reward for one
coffee, as the reward for the second coffee will be conditioned on having had a
first coffee already.}. This is the additivity requirement of the reward
function. The last requirement that we impose for the reward function should
capture the intuition suggested in the introduction, namely that more desirable
events should also be more probable events given the expectations of the
system. This is the consistency requirement.

We start out from a \emph{probability space} $(\Omega, \salgebra, \prob)$,
where $\Omega$ is the sample space, $\salgebra$ is a dense $\sigma$-algebra and
$\prob$ is a probability measure over $\salgebra$. In this section, we use
lowercase letters like $x,y,z$ to denote the elements of the $\sigma$-algebra
$\salgebra$. Given a set $\fs{X}$, its complement is denote by complement
$\fs{X}^\complement$ and its powerset by $\power(\fs{X})$.
The three desiderata can then be summarized as follows:

\begin{definition}[Reward]
Let $S = (\Omega, \salgebra, \prob)$ be a probability space. A function $\rew$
is a \emph{reward function} for $S$ iff it has the following three properties:
\begin{enumerate}
    \item \emph{Real-valued:} for all $x,y \in \salgebra$,
    \[
        \rew(x|y) \in \reals;
    \]
    \item \emph{Additivity:} for all $x,y,z \in \salgebra$,
    \[
        \rew(x,y|z) = \rew(x|z) + \rew(y|x,z);
    \]
    \item \emph{Consistent:} for all $x, y, u, v \in \salgebra$,
    \[
    \prob(x|u) > \prob(y|v) \quad\Longleftrightarrow\quad
        \rew(x|u) > \rew(y|v).
    \]
\end{enumerate}
Furthermore, the unconditional reward is defined as $\rew(x) \define
\rew(x|\Omega)$ for all $x \in \salgebra$.
\end{definition}

The following theorem shows that these three desiderata enforce a strict
mapping rewards and probabilities. The only function that can express such a
relationship is the logarithm.

\begin{theorem}\label{theo:reward} Let $S = (\Omega, \salgebra, \prob)$ be
a probability space. Then, a function $\rew$ is a reward function for $S$ iff
for all $x,y \in \salgebra$
\[
    \rew(x|y) = k \log \prob(x|y),
\]
where $k > 0$ is an arbitrary constant.
\end{theorem}

Notice that the constant $k$ in the expression $\rew(x|y) = k \log \prob(x|y)$
merely determines the units in which we choose to measure rewards. Thus, the
reward function $\rew$ for a probability space $(\Omega, \salgebra, \prob)$ is
essentially unique. As a convention, we will assume natural logarithms and set
the constant to $k = 1$, i.e. $\rew(x|y) = \ln \prob(x|y)$.

This result establishes a connection to information theory. It is immediately
clear that the reward of an event is nothing more than its negative information
content: the quantity $h(x) = -\rew(x)$ is the Shannon information content of
$x \in \salgebra$ measured in \emph{nats} \citep{MacKay2003}. This means that
we can interpret rewards as ``negative surprise values'', and that ``surprise
values'' constitute losses.

\begin{proposition}\label{prop:reward}
Let $\rew$ be a reward function over a probability space $(\Omega, \salgebra,
\prob)$. Then, it has the following properties:
\begin{itemize}
    \item[i.] Let $x,y \in \salgebra$. Then
        \[
            -\infty = \rew(\emptyset)
            \leq \rew(x|y) \leq
            \rew(\Omega) = 0.
        \]
    \item[ii.] Let $x \in \salgebra$ be an event. Then,
        \[
            e^{\rew(x^\complement)} = 1 - e^{\rew(x)}.
        \]
    \item[iii.] Let $z_1, z_2, \ldots \in \fs{F}$ be a sequence of
        disjoint events with rewards $\rew(z_1), \rew(z_2), \ldots$ and let
        $x = \bigcup_i z_i$. Then
        \[
            e^{\rew(x)} = \sum_i e^{\rew(z_i)}.
        \]
\end{itemize}
\end{proposition}

The proof of this proposition is trivial and left to the reader. The first part
sets the bounds for the values of rewards, and the two latter explain how to
construct the rewards of events from known rewards using complement and
countable union of disjoint events.

At a first glance, the fact that rewards take on complicated non-positive
values might seem unnatural, as in many applications one would like to use
numerical values drawn from arbitrary real intervals. Fortunately, given
numerical values representing the desirabilities of events, there is always an
affine transformation that converts them into rewards.

\begin{theorem}\label{theo:affine-transformation}
Let $\Omega$ be a countable set, and let $d: \Omega \rightarrow (-\infty, a]$
be a mapping. Then, for every $\alpha > 0$, there is a probability space
$(\Omega, \power(\Omega), \prob)$ with reward function $\rew$ such that:
\begin{enumerate}
    \item for all $\omega \in \Omega$,
    \[
        \rew(\{\omega\}) \define \alpha d(\omega) + \beta,
    \]
    where $\beta \define -\ln\left( \sum_{\omega' \in \Omega} e^{\alpha d(\omega')} \right)$;
    \item and for all $\omega, \omega' \in \Omega$,
    \[
        d(\omega) > d(\omega') \Leftrightarrow \rew(\{\omega\}) > \rew(\{\omega'\}).
    \]
\end{enumerate}
\end{theorem}

Note that Theorem~\ref{theo:affine-transformation} implies that the probability
$\prob(x)$ of any event $x$ in the $\sigma$-algebra $\power(\Omega)$ generated
by $\Omega$ is given by
\[
    \prob(x) =
    \frac{ \sum_{\omega \in x} e^{\alpha d(\omega)} }
         { \sum_{\omega \in \Omega} e^{\alpha d(\omega)} }.
\]
Note that for singletons $\{\omega\}$, $\prob(\{\omega\})$ is the Gibbs measure
with negative energy $d(\omega)$ and temperature $\propto \frac{1}{\alpha}$. It
is due to this analogy that we call the quantity $\frac{1}{\alpha} > 0$ the
\emph{temperature parameter} of the transformation.

\section{Utilities in Stochastic Processes}\label{sec:Utility}

In this section, we consider a \emph{stochastic process} $\prob$ over sequences
$x_1 x_2 x_3 \cdots$ in $\fs{X}^\infty$. We specify the process by
assigning conditional probabilities $\prob(x_t|x_{<t})$ to all finite strings
$x_{\leq t} \in \fs{X}^\ast$. Note that the distribution $\prob(x_{\leq t}) = \prod_{\tau=1}^t
\prob(x_\tau|x_{<\tau})$ for all $x_{\leq t} \in \fs{X}^\ast$ is normalized
by construction. By the
Kolmogorov extension theorem, it is guaranteed that there exists a unique probability
space $S = (\fs{X}^\infty, \salgebra, \prob)$. We therefore omit the reference
to $S$ and talk about the process $\prob$.

The reward function $\rew$ derived in the previous section correctly expresses
preference relations amongst different outcomes. However, in the context of
random sequences, it has the downside that the reward of most sequences
diverges. A sequence $x_1 x_2 x_3 \cdots$ can be interpreted as a progressive
refinement of a point event in $\salgebra$, namely, the sequence of events
$\epsilon \supset x_{\leq 1} \supset x_{\leq 2} \supset x_{\leq 3} \supset
\cdots$. One can exploit the interpretation of the index as time to define a
quantity that does not diverge. We define thus the utility as the reward rate
of a sequence.

\begin{definition}[Utility]
\label{def:utility} Let $\rew$ be a reward function for the process $\prob$.
The utility of a string $x_{\leq t} \in \fs{X}^\ast$ is defined as
\[
    \utility(x_{\leq t}) \define \frac{1}{t} \sum_{\tau=1}^t
    \rew(x_\tau|x_{<\tau}),
\]
and for a sequence $x = x_1 x_2 x_3 \cdots \in \fs{X}^\infty$ it is defined as
\[
    \utility(x) \define \lim_{t \rightarrow \infty}
        \utility(x_{\leq t})
\]
if this limit exists\footnote{Strictly speaking, one could define the upper and
lower rate $\utility^{+}(x) \define \limsup_{t \rightarrow \infty}
\utility(x_{\leq t})$ and $\utility^{-}(x) \define \liminf_{t \rightarrow
\infty} \utility(x_{\leq t})$ respectively, but we avoid this distinction for
simplicity.}.
\end{definition}

A utility function that is constructed according to
Definition~\ref{def:utility} has the following properties.

\begin{proposition}\label{prop:utility}
Let $\utility$ be a utility function for a process $\prob$. The following properties
hold:
\begin{itemize}
    \item[i.] For all $x = x_1 x_2 \cdots \in \fs{X}^\infty$ and all $t, k \in \nats$,
        \[
            -\infty = \utility(\lambda)
            \leq \utility(x_{\leq t}) \leq
            \utility(\epsilon) = 0,
        \]
        where $\lambda$ is any impossible string/sequence.
    \item[ii.] For all $x_{\leq t} \in \fs{X}^\ast$,
        \[
            \prob(x_{\leq t})
            = \exp\Bigl( t \cdot \utility(x_{\leq t}) \Bigr).
        \]
    \item[iii.] For any $t \in \nats$,
        \[
                \expect[\utility(x_{\leq t})]
                = -\frac{1}{t}\entropy[ \prob(x_{\leq t}) ],
        \]
        where $\entropy$ is the entropy functional (see the appendix).
\end{itemize}
\end{proposition}
Part~(i) provides trivial bounds on the utilities that directly carry over from
the bounds on rewards. Part~(ii) shows how the utility of a sequence determines
its probability. Part~(iii) implies that the expected utility of an interaction
sequence is just its negative entropy rate.

\section{Utility in Coupled I/O systems}

Let $\fs{O}$ and $\fs{A}$ be two finite sets, the first being the \emph{set of
observations} and the second being the \emph{set of actions}. Using $\fs{A}$
and $\fs{O}$, a set of interaction sequences is constructed. Define the
\emph{set of interactions} as $\fs{Z} \define \fs{A} \times \fs{O}$. A pair
$(a,o) \in \fs{Z}$ is called an \emph{interaction}. We underline symbols to
glue them together as in $\g{ao}_{\leq t} = a_1 o_1 a_2 o_2 \cdots a_t o_t$.

An \emph{I/O system} $\prob$ is a probability distribution over interaction
sequences $\infint$. $\prob$ is uniquely determined by the conditional
probabilities
\[
    \prob(a_t|\g{ao}_{<t}), \quad \prob(o_t|\g{ao}_{<t}a_t)
\]
for each $\g{ao}_{\leq t} \in \finint$. However, the semantics of the
probability distribution $\prob$ are only fully defined once it is coupled to
another system. Note that an I/O system is formally equivalent to a stochastic
process; hence one can construct a reward function $\rew$ for $\prob$.

Let $\agent$, $\env$ be two I/O systems. An \emph{interaction system} $(\agent,
\env)$ defines a \emph{generative
distribution} $\gen$ that describes the probabilities that actually govern the
I/O stream once the two systems are coupled. $\gen$ is specified by the
equations
\begin{align*}
    \gen(a_t|\g{ao}_{<t}) &= \agent(a_t|\g{ao}_{<t}) \\
    \gen(o_t|\g{ao}_{<t}a_t) &= \env(o_t|\g{ao}_{<t}a_t)
\end{align*}
valid for all $\g{ao}_t \in \finint$. Here, $\gen$ is a stochastic process over
$\infint$ that models the true probability distribution over interaction
sequences that arises by coupling two systems through their I/O streams. More
specifically, for the system $\agent$, $\agent(a_t|\g{ao}_{<t})$ is the
probability of producing action $a_t \in \fs{A}$ given history $\g{ao}_{<t}$
and $\agent(o_t|\g{ao}_{<t}a_t)$ is the predicted probability of the
observation $o_t \in \fs{O}$ given history $\g{ao}_{<t}a_t$. Hence, for
$\agent$, the sequence $o_1 o_2 \ldots$ is its input stream and the sequence
$a_1 a_2 \ldots$ is its output stream. In contrast, the roles of actions and
observations are reversed in the case of the system $\env$.
This model of interaction is very general in that it can
accommodate many specific regimes of interaction. By convention, we call the
system $\agent$ the \emph{agent} and the system $\env$ the \emph{environment}.

In the following we are interested in
understanding the actual utilities that can be achieved by an agent
$\agent$ once coupled to a particular environment $\env$.
Accordingly, we will compute expectations over functions of interaction
sequences with respect to $\gen$, since the generative distribution $\gen$ describes
the actual interaction statistics of the two coupled I/O systems.

\begin{theorem}\label{theo:exp-reward}
Let $(\agent,\env)$ be an interaction system. The expected rewards of $\gen$,
$\agent$ and $\env$ for the first $t$ interactions are given by
\begin{align*}
    \expect[\rew_\gen(\g{ao}_{\leq t})] =
    &- \entropy[\agent(a_{\leq t}|o_{<t})]
    - \entropy[\env(o_{\leq t}|a_{\leq t})],
    \\
    \expect[\rew_\agent(\g{ao}_{\leq t})] =
    &- \entropy[\agent(a_{\leq t}|o_{<t})]
    - \entropy[\env(o_{\leq t}|a_{\leq t})] \\
    &- \kldiv[\env(o_{\leq t}|a_{\leq t}) \| \agent(o_{\leq t}|a_{\leq t})],
    \\
    \expect[\rew_\env(\g{ao}_{\leq t})] =
    &- \entropy[\agent(a_{\leq t}|o_{<t})]
    - \entropy[\env(o_{\leq t}|a_{\leq t})] \\
    &- \kldiv[\agent(a_{\leq t}|o_{<t}) \| \env(a_{\leq t}|o_{<t})],
\end{align*}
where $\rew_\gen$, $\rew_\agent$ and $\rew_\env$ are the reward functions for
$\gen$, $\agent$ and $\env$ respectively. Note that $\entropy$ and
$\kldiv$ are the entropy and the relative entropy functionals as defined in the
appendix.
\end{theorem}

Accordingly, the interaction system's expected reward is given by the negative
sum of the entropies produced by the agent's action generation probabilities
and the environment's observation generation probabilities. The agent's
(actual) expected reward is given by the negative cross-entropy between the
generative distribution $\gen$ and the agent's distribution $\agent$. The
discrepancy between the agent's and the interaction system's expected reward is
given by the relative entropy between the two probability distributions. Since
the relative entropy is positive, one has $\expect[\rew_\gen(\g{ao}_{\leq t})]
\geq \expect[\rew_\agent(\g{ao}_{\leq t})]$. This term implies that the better
the environment is ``modeled'' by the agent, the better its performance will
be. In other words: the agent has to recognize the structure of the environment
to be able to exploit it. The designer can directly increase the agent's
expected performance by controlling the first and the last term. The middle
term is determined by the environment and only indirectly controllable.
Importantly, the terms are in general coupled and not independent: changing one
might affect another. For example, the first term suggests that less stochastic
policies improve performance, which is oftentimes the case. However, in the
case of a game with mixed Nash equilibria the overall reward
can increase for a stochastic policy, which means that the first term is
compensated for by the third term. Given the expected rewards, we can easily
calculate the expected utilities in terms of entropy rates.

\begin{corollary}\label{cor:expected-utility}
Let $(\agent,\env)$ be an interaction system. The expected utilities of $\gen$,
$\agent$ and $\env$ are given by
\begin{align*}
    \expect[\utility_\gen]
    &= \genutility_\agent + \genutility_\env
    \\
    \expect[\utility_\agent]
    &= \genutility_\agent + \genutility_\env
        + \predutility_\agent
    \\
    \expect[\utility_\env]
    &= \genutility_\agent + \genutility_\env
        + \predutility_\env
\end{align*}
where $\genutility_\agent$, $\genutility_\env$ and $\predutility_\agent$ are
entropy rates defined as
\begin{small}
\begin{align*}
    \genutility_\agent &\define
        -\frac{1}{t} \sum_{\tau=1}^t
        \entropy[\agent(a_\tau|\g{ao}_{<\tau})]
    \\
    \predutility_\agent &\define
        -\frac{1}{t}\sum_{\tau=1}^t
        \kldiv[\env(o_\tau|\g{ao}_{<\tau}a_\tau)
            \| \agent(o_\tau|\g{ao}_{<\tau}a_\tau)]
    \\
    \genutility_\env &\define
        -\frac{1}{t} \sum_{\tau=1}^t
        \entropy[\env(o_\tau|\g{ao}_{<\tau}a_\tau)]
    \\
    \predutility_\env &\define
         -\frac{1}{t}\sum_{\tau=1}^t
         \kldiv[\agent(a_\tau|\g{ao}_{<\tau})
            \| \env(a_\tau|\g{ao}_{<\tau})].
\end{align*}
\end{small}
\end{corollary}

This result is easily obtained by dividing the quantities in
Theorem~\ref{theo:exp-reward} by $t$ and then applying the chain rule for
entropies to break the rewards over full sequences into instantaneous rewards.
Note that $\genutility_\agent$, $\genutility_\env$ are the contributions to the
utility due the generation of interactions, and $\predutility_\agent$,
$\predutility_\env$ are the contributions to the utility due to the prediction
of interactions.

\section{Examples}

\begin{figure*}
\centering
    \footnotesize
    \includegraphics[scale=0.6]{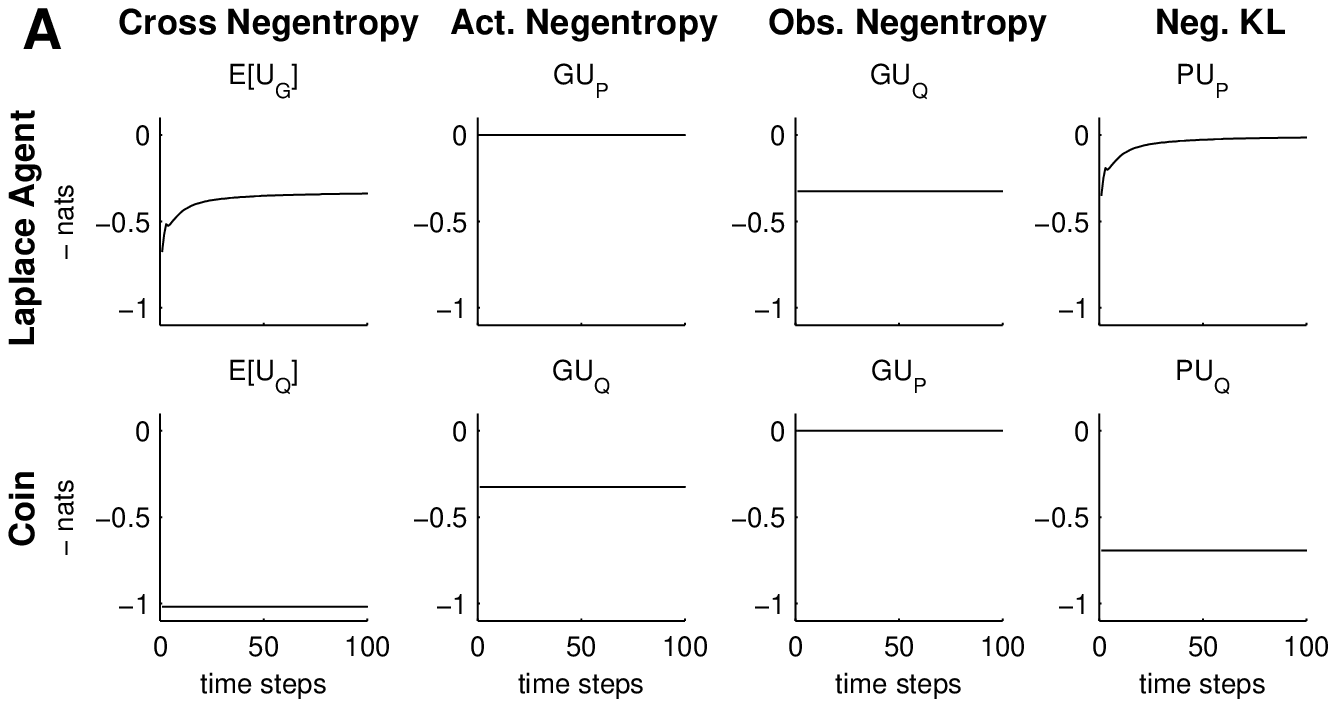}
    \includegraphics[scale=0.6]{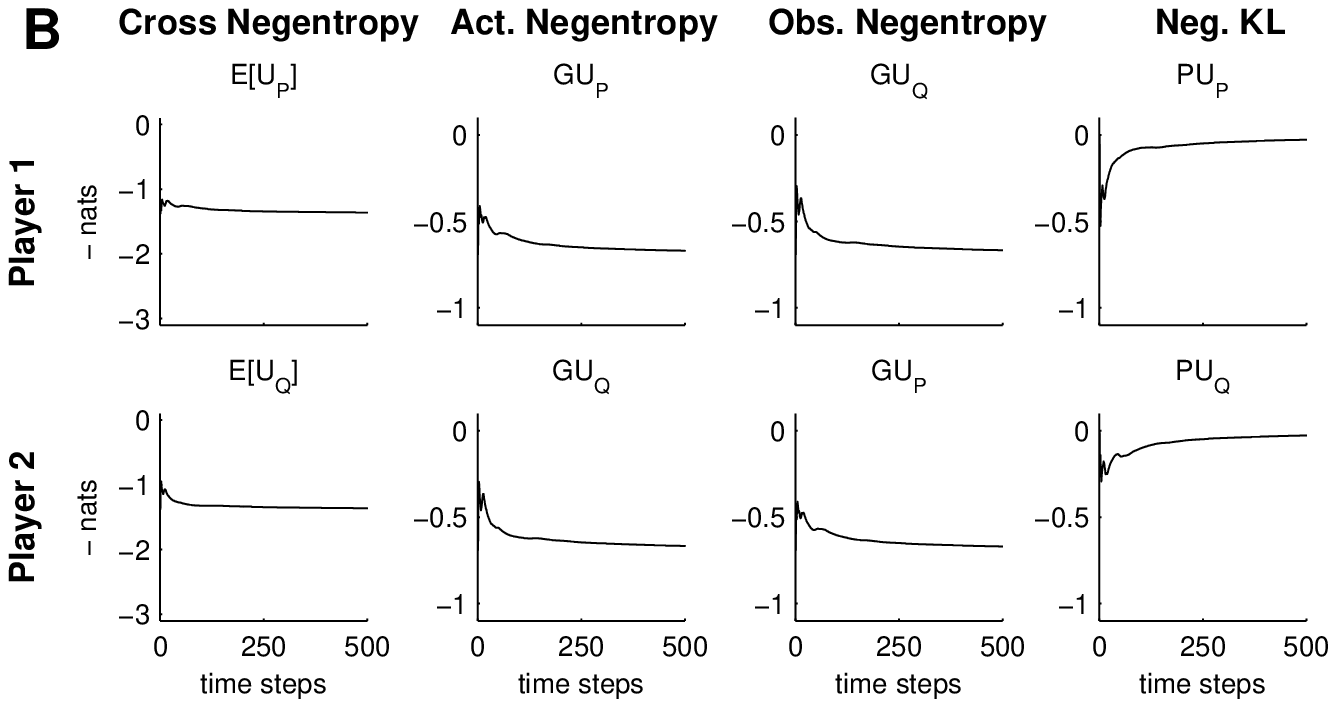}
    \caption{
    (A) Entropy dynamics of a Laplace agent interacting with a coin of bias $0.9$. The Laplace agent learns
    to predict the coin's behavior as can be seen in the decrease of the KL-divergence and the cross
    entropy. Since the Laplace agent acts deterministically its action entropy is always zero. Its
    observation entropy equals the action entropy of the coin. The coin does not change its behavior,
    which can be seen from the flat entropy curves. (B) Entropy dynamics of two adaptive agents playing matching pennies. Both agents follow
    smooth fictitious play. They converge to uniform random policies, which means that their
    action negentropies converge to $\log(2)$. Both agents learn the probability distribution
    of the other agent, as can be seen in the decrease of the KL-divergences.
    }\label{fig:matching-pennies}
\end{figure*}

One of the most interesting aspects of the information-theoretic formulation of
utility is that it can be applied both to control problems (where an agent acts
in a non-adaptive environment) and to game theoretic problems (where two
possibly adaptive agents interact). In the following we apply the proposed
utility measures to two simple toy examples from these two areas. In the first
example, an adaptive agent interacts with a biased coin (the non-adaptive
agent) and tries to predict the next outcome of the coin toss, which is either
`Head' (H) or `Tail' (T). In the second example two adaptive agents interact
playing the \emph{matching pennies} game. One player has to match her action
with the other player (HH or TT), while the other player has to unmatch (TH or
HT). All agents have the same sets of possible observations and actions which
are the binary sets $\fs{O}=\{\mathrm{H},\mathrm{T}\}$ and
$\fs{A}=\{\mathrm{H},\mathrm{T}\}$.

\paragraph{Example 1.}
The non-adaptive agent is a biased coin. Accordingly, the coin's action
probability is given by its bias and was set to $\env(o=\mathrm{H})=0.9$. The
coin does not have any biased expectations about its observations, so we set
$\env(a=\mathrm{H})=0.5$. The adaptive agent is given by the Laplace agent
whose expectations over observed coin tosses follows the predictive
distribution $\agent(o=\mathrm{H}|t,n)=(n+1)/(t+2)$, where $t$ is the number of
coin tosses observed so far and $n$ is the number of observed Heads. Based on
this estimator the Laplace agent chooses its action deterministically according
to $\agent(a=\mathrm{H}|t,n)=\Theta(\frac{n+1}{t+2}-\frac{1}{2})$, where
$\Theta(\cdot)$ is the Heaviside step function. From these distributions the
full probability over interaction sequences can be computed. Figure~1A shows
the entropy dynamics for a typical single run. The Laplace agent learns the
distribution of the coin tosses, i.e. the KL decreases to zero. The negative
cross-entropy stabilizes at the value of the observation entropy that cannot be
further reduced. The entropy dynamics of the coin do not show any modulation.

\paragraph{Example 2.}
The two agents are modeled based on \emph{smooth fictitious play}
\citep{FudenbergKreps1993}. Both players keep count of the empirical
frequencies of Head and Tail respectively. Therefore, each player $i$ stores
the quantities $\kappa_i^{(1)} = n_i$ and $\kappa_i^{(2)} = t-n_i$ where $t$ is
the number of moves observed so far, $n_1$ is the number of Heads observed by
Player~1 and $n_2$ is the number of Heads observed by Player~2. The probability
distributions $\agent(o=\mathrm{H}|t,n_1)=\gamma_1$ and
$\env(a=\mathrm{H}|t,n_2)=\gamma_2$ over inputs is given by these empirical
frequencies through $\gamma_i = \kappa_i / \sum_i \kappa_i$. The action
probabilities are computed according to a sigmoid best-response function
$\agent(a=\mathrm{H}|t,n_1)=1 / (1+\exp(-\alpha (\gamma_1-0.5) ))$, and
$\env(o=\mathrm{H}|t,n_2)=1 / (1+\exp(-\alpha (0.5-\gamma_2) ))$ respectively
in case of Player~2 that has to unmatch. This game has a well-known equilibrium
solution that is a mixed strategy Nash equilibrium where both players act
randomly. Both action and observation entropies converge to the value
$\log(2)$. Interestingly, the information-theoretic utility as computed by the
cross-entropy takes the action entropy into account. Compare Figure~1B.

\section{Conclusion}%

Based on three simple desiderata we propose that rewards can be measured
in terms of information content and that, consequently, the entropy
satisfies properties characteristic of a utility function.
Previous theoretical studies have reported structural similarities between
entropy and utility functions, see e.g. \citep{Candeal2001}, and
recently, relative entropy has even been proposed as a measure of utility
in control systems \citep{Todorov2009,Kappen2009,OrtegaBraun2008}.
The contribution of this paper
is to derive axiomatically a precise relation between rewards
and information value and to apply it to coupled I/O systems.

The utility functions that we have derived can be
conceptualized as \emph{path utilities}, because they
assign a utility value to an entire history.
This is very similar to the path integral formulation in
quantum mechanics where the utility of a path is determined
by the classic action integral and the probability of a path
is also obtain by taking the exponential of this `utility' \citep{Feynman1965}.
In particular, we obtain the (cumulative time-averaged)
cross entropy as a utility function
when an agent is coupled to an environment. This utility function
not only takes into account the KL-divergence as a measure
of learning, but also the action entropy. This is interesting,
because in most control problems
controllers are designed to be deterministic (e.g. optimal control theory)
in response to a known and stationary environment.
If, however, the environment is not stationary and in fact adaptive
as well, then it is a well-known result from game theory that
optimal strategies might be randomized. The utility function
that we are proposing might indeed allow quantifying
a trade-off between reducing the KL and reducing the
action entropy. In the future it will therefore be
interesting to investigate this utility function in more complex
interaction systems.


\appendix

\section{Appendix}

\subsection{Entropy functionals}
\emph{Entropy:} Let $\prob$ be a probability distribution over \mbox{$\fs{X}
\times \fs{Y}$}. Define the (average conditional) entropy \citep{Shannon1948}
as
\[
    \entropy[ \prob(x|y) ]
    \define -\sum_{x,y} \prob(x,y) \ln \prob(x|y).
\]
\emph{Relative Entropy:} Let $\prob_1$ and $\prob_2$ be two probability
distributions over $\fs{X} \times \fs{Y}$. Define the (average conditional)
relative entropy \citep{KullbackLeibler1951} as
\[
    \kldiv[ \prob_1(x|y) \| \prob_2(x|y) ]
    \define \sum_{x,y} \prob_1(x,y)
        \ln \frac{ \prob_1(x|y) }{ \prob_2(x|y) }.
\]

\subsection{Proof of Theorem~\ref{theo:reward}\label{proof:reward}}

\begin{proof}
Let the function $g$ be such that $g(\prob(x)) = \rew(x)$. Let $x_1, x_2,
\ldots, x_n \in \salgebra$ be a sequence of events, such that $\prob(x_1) =
\prob(x_i|x_{<i}) > 0$ for all $i = 2,\ldots,n$. We have $\prob(x_1,\ldots,x_n)
= \prod_i \prob(x_i|x_{<i}) = \prob(x_1)^n$. Since $\prob(x) > \prob(x')
\Leftrightarrow \rew(x) > \rew(x')$ for any $x, x' \in \salgebra$, then
$\prob(x) = \prob(x') \Leftrightarrow \rew(x) = \rew(x')$, and thus $\prob(x_1)
= \prob(x_i|x_{<i}) \Leftrightarrow \rew(x_1) = \rew(x_i|x_{<i})$ for all
$i=2,\ldots,n$. This means, $\rew(x_1,\ldots,x_n) = n \rew(x_i)$. But
$g(\prob(x_1,\ldots,x_n)) = \rew(x_1,\ldots,x_n)$, and hence $g(\prob(x_1)^n) =
n \rew(x_1)$. Similarly, for a second sequence of events $y_1, y_2, \ldots, y_m
\in \fs{F}$ with $\prob(y_1) = \prob(y_i|y_{<i}) > 0$ for all $i=1,\ldots,m$,
we have $g(\prob(y_1)^n) = n \rew(y_1)$.

The rest of the argument parallels Shannon's entropy theorem
\citep{Shannon1948}. Define $p = \prob(x_1)$ and $q = \prob(y_1)$. Choose $n$
arbitrarily high to satisfy $q^m \leq p^n < q^{m+1}$. Taking the logarithm, and
dividing by $n \log q$ one obtains
\[
    \frac{m}{n} \leq \frac{\log p}{\log q} < \frac{m}{n} + \frac{1}{n}
    \quad\Leftrightarrow\quad
    \Bigl| \frac{m}{n} - \frac{\log p}{\log q} \Bigr| < \varepsilon,
\]
where $\varepsilon > 0$ is arbitrarily small. Similarly, using
$g(p^n) = n \, g(p)$ and the monotonicity of $g$, we can write
$m \, g(q) \leq n \, g(p) < (m+1) \, g(q)$ and thus
\[
    \frac{m}{n} \leq \frac{g(p)}{g(q)} < \frac{m}{n} + \frac{1}{n}
    \quad\Leftrightarrow\quad
    \Bigl| \frac{m}{n} - \frac{g(p)}{g(q)} \Bigr| < \varepsilon,
\]
where $\varepsilon > 0$ is arbitrarily small. Combining these two
inequalities, one gets
\[
    \Bigl| \frac{\log p}{\log q} - \frac{g(p)}{g(q)} \Bigr| < 2\varepsilon,
\]
which, fixing $q$, gives $\rew(p) = g(p) = k \log p$, where $k > 0$. This holds
for any $x_1 \in \fs{F}$ with $\prob(x_1) > 0$.
\end{proof}

\subsection{Proof of Theorem~\ref{theo:affine-transformation}}
\begin{proof}
For all $\omega, \omega' \in \Omega$, $d(\omega) > d(\omega') \Leftrightarrow
\alpha d(\omega) + \beta > \alpha d(\omega') + \beta \Leftrightarrow
\rew(\{\omega\}) > \rew(\{\omega'\})$ because the affine transformation is
positive. Now, the induced probability over $\power(\Omega)$ has atoms
$\{\omega\}$ with probabilities $\prob(\{\omega\}) = e^{\rew(\{\omega)\}} \geq
0$ and is normalized:
\[
    \sum_{\omega \in \Omega} e^{\rew(\{\omega\})}
    = \sum_{\omega \in \Omega} e^{\alpha d(\{\omega\}) + \beta}
    = \frac{ \sum_{\omega \in \Omega} e^{\alpha d(\omega)} }
           { \sum_{\omega \in \Omega} e^{\alpha d(\omega)} }
    = 1.
\]
Since knowing $\prob(\{\omega\})$ for all $\omega \in \Omega$ determines the
measure for the whole field $\power(\Omega)$, $(\Omega,\power(\Omega),\prob)$
is a probability space.
\end{proof}

\subsection{Proof of Proposition~\ref{prop:utility}}

\begin{proof}
(i) Since $-\infty < \rew(x_\tau|x_{<\tau}) \leq 0$ for all $\tau$, then
$-\infty < \frac{1}{t}\sum_{\tau=1}^t\rew(x_\tau|x_{<\tau}) = \utility(x_{\leq
t}) \leq 0$ for all $t$. (ii) Write $\prob(x_{\leq t})$ as
\begin{align*}
    \prob(x_{\leq t})
    &= \prod_{\tau=1}^t \prob(x_\tau|x_{<\tau})
    = \prod_{\tau=1}^t \exp\Bigl( \rew(x_\tau|x_{<\tau}) \Bigr) \\
    &= \exp\Bigl( \sum_{\tau=1}^t \rew(x_\tau|x_{<\tau}) \Bigr)
    = \exp\Bigl( t \cdot \utility(x_{\leq t}) \Bigr).
\end{align*}
(iii) $\expect[\utility(x_{\leq t})] = \sum_{x_{\leq t}} \prob(x_{\leq t})
\utility(x_{\leq t}) = \sum_{x_{\leq t}} \prob(x_{\leq t})
\frac{1}{t}\rew(x_{\leq t}) = -\frac{1}{t} \entropy[ \prob(x_{\leq t}) ]$,
where we have applied~(ii) in the second equality and $\rew(\cdot) =
\ln(\prob(\cdot))$ in the third equality.
\end{proof}

\subsection{Proof of Theorem~\ref{theo:exp-reward}}
\begin{proof}
This proof is done by straightforward calculation. First note that
\begin{align*}
    \gen(\g{ao}_{\leq t})
    &= \prod_{\tau=1}^t \agent(a_\tau|\g{ao}_{<\tau}) \env(o_\tau|\g{ao}_{<\tau}a_\tau) \\
    &= \agent(a_{\leq t}|o_{<t}) \env(o_{\leq t}|a_{\leq t}),
\end{align*}
which is obtained by applying multiple times the chain rule for probabilities
and noting that the probability of a symbol is fully determined by the previous
symbols. Similarly $\agent(\g{ao}_{\leq t}) = \agent(a_{\leq t}|o_{<t})
\agent(o_{\leq t}|a_{\leq t})$ is obtained. We calculate here
$\expect[\rew_\agent(\g{ao}_{\leq t})]$. The calculation for
$\expect[\rew_\gen(\g{ao}_{\leq t})]$ and $\expect[\rew_\env(\g{ao}_{\leq t})]$
are omitted because they are analogous.
{\small
\begin{align*}
    \expect[\rew_\agent(\g{ao}_{\leq t})]
    &\stackrel{(a)}{=}
        \sum_{\g{ao}_{\leq t}} \gen(\g{ao}_{\leq t})
            \ln \agent(\g{ao}_{\leq t}) \\
    &\stackrel{(b)}{=}
        \sum_{\g{ao}_{\leq t}}
        \gen(\g{ao}_{\leq t})
        \Bigl(
            \ln \agent(a_{\leq t}|o_{<t}) + \ln \agent(o_{\leq t}|a_{\leq t})
        \Bigr)
    \\&\stackrel{(c)}{=}
        \sum_{\g{ao}_{\leq t}}
        \gen(\g{ao}_{\leq t})
        \Bigl(
            \ln \agent(a_{\leq t}|o_{<t}) + \ln \agent(o_{\leq t}|a_{\leq t}) \\
            &\qquad + \ln \env(o_{\leq t}|a_{\leq t}) - \ln \env(o_{\leq t}|a_{\leq t})
        \Bigr)
    \\&\stackrel{(d)}{=}
        - \entropy[\agent(a_{\leq t}|o_{<t})]
        - \entropy[\env(o_{\leq t}|a_{\leq t})]
        \\&\qquad- \kldiv[\env(o_{\leq t}|a_{\leq t}) \| \agent(o_{\leq t}|a_{\leq t})].
\end{align*}
}
Equality~(a) follows from the definition of expectations and the relation
between rewards and probabilities. In~(b) we separate the term in the logarithm
into the action and observation part. In~(c) we add and subtract the term
$\env(o_{\leq t}|a_{\leq t})$ in the logarithm. Equality~(d) follows from the
algebraic manipulation of the terms and from identifying the entropy terms,
noting that $\gen(\g{ao}_{\leq t}) = \agent(a_{\leq t}|o_{<t}) \env(o_{\leq
t}|a_{\leq t})$.
\end{proof}

\vskip 0.2in
\bibliographystyle{unsrt}
\footnotesize
\bibliography{bibliography}

\end{document}